\documentclass{article}
\usepackage{spconf,amsmath,graphicx}
\usepackage{times}
\usepackage{latexsym}
\usepackage{amsmath}
\usepackage{algorithm}
\usepackage{algpseudocode}
\usepackage{url}
\usepackage{multirow}
\usepackage{CJKutf8} 
\usepackage{ textcomp }
\usepackage{comment} 
\usepackage{amsfonts}

\title{ODSQA: OPEN-DOMAIN SPOKEN QUESTION ANSWERING DATASET}
\name{Chia-Hsuan Lee, Shang-Ming Wang, Huan-Cheng Chang, Hung-Yi Lee\thanks{Thanks to Delta Research Center
and Delta Electronics, Inc. for collecting the DRCD dataset.}}
\address{ College of Electrical Engineering and Computer Science, National Taiwan University, Taiwan\\}
\begin{document}
%\ninept
%
\maketitle
\begin{abstract}
Reading comprehension by machine has been widely studied, but machine comprehension of spoken content is still a less investigated problem. In this paper, we release Open-Domain Spoken Question Answering Dataset (ODSQA) with more than three thousand questions. To the best of our knowledge, this is the largest real SQA dataset. On this dataset, we found that ASR errors have catastrophic impact on SQA.
To mitigate the effect of ASR errors, subword units are involved, which brings consistent improvements over all the models. We further found that data augmentation on text-based QA training examples can improve SQA. 

\end{abstract}
\section{Introduction}
\label{sec:intro}

Machine comprehension and question answering on text have significant progress in the recent years.
One of the most representative corpora is the Stanford Question Answering Dataset (SQuAD)~\cite{rajpurkar2016squad}, on which deep neural network- (DNN-) based models are  comparable with human. %Exact Match(EM) score 82.3 with 82.48. 
The achievements of the state-of-the-art question answering models demonstrate that machine has already acquired complex reasoning ability.
On the other hand, accessing large collections of  multimedia or spoken content is much more difficult and time-consuming than  plain text content for humans. 
It is therefore highly attractive to develop Spoken Question Answering (SQA)~\cite{ispoken,comas2012factoid,turmo2008overview,comas2012sibyl}, which requires machine to find the answer from spoken content given a question in either text or spoken form. 

In SQA, after transcribing spoken content into text by automatic speech recognition (ASR), typical approaches use information retrieval (IR) techniques~\cite{shiang2014spoken} or  knowledge bases~\cite{hixon2015learning} to find the proper answer from the transcriptions. 
Another attempt towards machine comprehension of spoken content is TOEFL listening comprehension by machine~\cite{tseng2016towards}. %in which the machine is given an audio story, and required to answer the text questions related to that audio story. 
TOEFL is an English examination that tests the knowledge and skills of academic English for English learners whose native languages are not English. 
Deep-based models including   attention-based RNN\cite{tseng2016towards} and tree-structured RNN\cite{fang2016hierarchical} were used to answer TOEFL listening comprehension test.
Transfer learning for Question Answering (QA) is also studied on this task\cite{chung2017supervised}. 
However, TOEFL listening comprehension test is a multi-select question answering corpus, and its scale is not large enough to support the training of powerful listening comprehension models. 
Another spoken question answering corpus is Spoken-SQuAD\cite{li2018spoken}, which is generated from SQuAD dataset through Google Text-to-Speech (TTS) system. The spoken content is then transcribed by CMU sphinx\cite{walker2004sphinx}. 
Several state-of-the-art question answering models are evaluated on this dataset, and ASR errors seriously  degrade the performance of these models. 
On Spoken-SQuAD, it has been verified that using sub-word units in SQA can  mitigate the impact of ASR errors.
Although Spoken-SQuAD is large enough to train state-of-the-art QA models, it is artificially generated, so it is still one step away from real SQA. 

To further push the boundary of SQA, in this paper, we release a large scale SQA dataset -- Open-Domain Spoken Question Answering Dataset (ODSQA). 
The contribution of our work are four-fold: 
\begin{itemize}
  \item First of all, we release an SQA dataset, ODSQA, with more than three thousand questions. 
  ODSQA is a Chinese dataset, and  to the best of our knowledge, the largest real\footnote{not generated by TTS as Spoken-SQA} SQA dataset for extraction-based QA task. 
  \item Secondly, we found ASR errors have catastrophic impact on real SQA. We tested numbers of state-of-the-art SQuAD models on ODSQA, and reported their degrading performance on ASR transcriptions.
  %\item Last but not the least, we introduce domain adversarial learning to improve the performance of spoken question answering by learning domain-general features between text and ASR transcriptions. Experimentally, we improve 000\% F1 score on a Chinese SQA corpus compared to the baseline model which is only trained with target language examples. 
  \item Thirdly, we apply sub-word units in SQA to mitigate the impact of speech recognition errors, and this approach brings consistent improvements experimentally.
  \item Last but not the least, we found that back-translation, which has been applied on text QA~\cite{yu2018qanet} to improve the performance of models,  also improve the SQA models. 
\end{itemize}

\section{Related Work}
% introduce QA datasets

Most QA work focuses on understanding text documents\cite{richardson2013mctest,lai2017race,rajpurkar2016squad,trischler2016newsqa}.
The QA task has been extended from text to  images
\cite{zitnick2016adopting,young2014image,lin2014microsoft,kong2014you} or video descriptions \cite{chen2011collecting,das2013thousand,rohrbach2015dataset}.
In the MovieQA task\cite{tapaswi2016movieqa}, the machine answers questions about movies using video clips, plots, subtitles, scripts, and DVS.
Usually only text information (e.g., the movie's plot) is considered in the MovieQA task; learning to answer questions using video is still difficult.
Machine comprehension of spoken content is still a less investigated problem.
%https://www.sharelatex.com/project/5accc180b0b9414dc05b1ce2

To mitigate the impact of speech recognition errors, we use sub-word units to represent the transcriptions of spoken content in this paper. 
Using sub-word unit is a popular approach for speech-related down-stream task and has been applied to spoken document retrieval\cite{ng1997subword}, spoken term detection \cite{van2017constructing}\cite{huijbregts2011unsupervised}, spoken document categorization\cite{qu2000using}, and speech recognition\cite{parada2011learning}. 
It has been verified that sub-word units can improve the performance of SQA~\cite{li2018spoken} .
However, the previous experiments only conducted on an artificial dataset.
In addition, the previous work focuses on English SQA, whereas we focus on Chinese SQA in this paper.
There is a big difference between the subword units of English and Chinese.
%have been applied on spoken question answering.
%The most related study to our work is \cite{spoken squad} in which phonetic sub-word units have been applied on spoken question answering. However, \cite{spoken squad} used sub-word units from English data. We focus on Chinese data in this work and there is a big difference between English syllable and Chinese syllable.

To improve the robustness to speech recognition errors, we used back-translation as a data augmentation approach in this paper. 
Back-translation allows the model to learn from  more diversified data through paraphrasing. 
Back-translation was also studied in spoken language understanding and text-based QA as a data augmentation approach. 
%################################# Lee:這一段看不懂
In cross lingual spoken language understanding, training with the back-translation data via target language will make the model adaptive to translation errors \cite{he2013multi}\cite{upadhyay2018almost}. 
%#################################
In text-based QA, back-translation was used to paraphrase questions\cite{dong2017learning} and paraphrase documents\cite{yu2018qanet}.  

%Transfer learning of QA between different datasets in the same language has been studied~\cite{kadlec2016particular,chung2017supervised,min2017question,wiese2017neural,golub2017two}.\
% introduce Transfer learning in QA (domain adversarial ?!?!!!!)
%Domain adaptation, or in a broader view, transfer learning has been studied for question answering. They mainly focus on pre-training on a larger scale dataset and adpat the model to a relatively smaller dataset in supervised \cite{min2017question,wiese2017neural,kadlec2016particular} or even unsupervised way \cite{chung2017supervised,golub2017two}. However, to the best of our knowledge, we are the first to apply transfer learning for ASR adaptations on question answering.

\section{Task Description}

\subsection{Data Format}
\label{sec:format}
\begin{figure}[t]
  \centering
  \includegraphics[width=\linewidth]{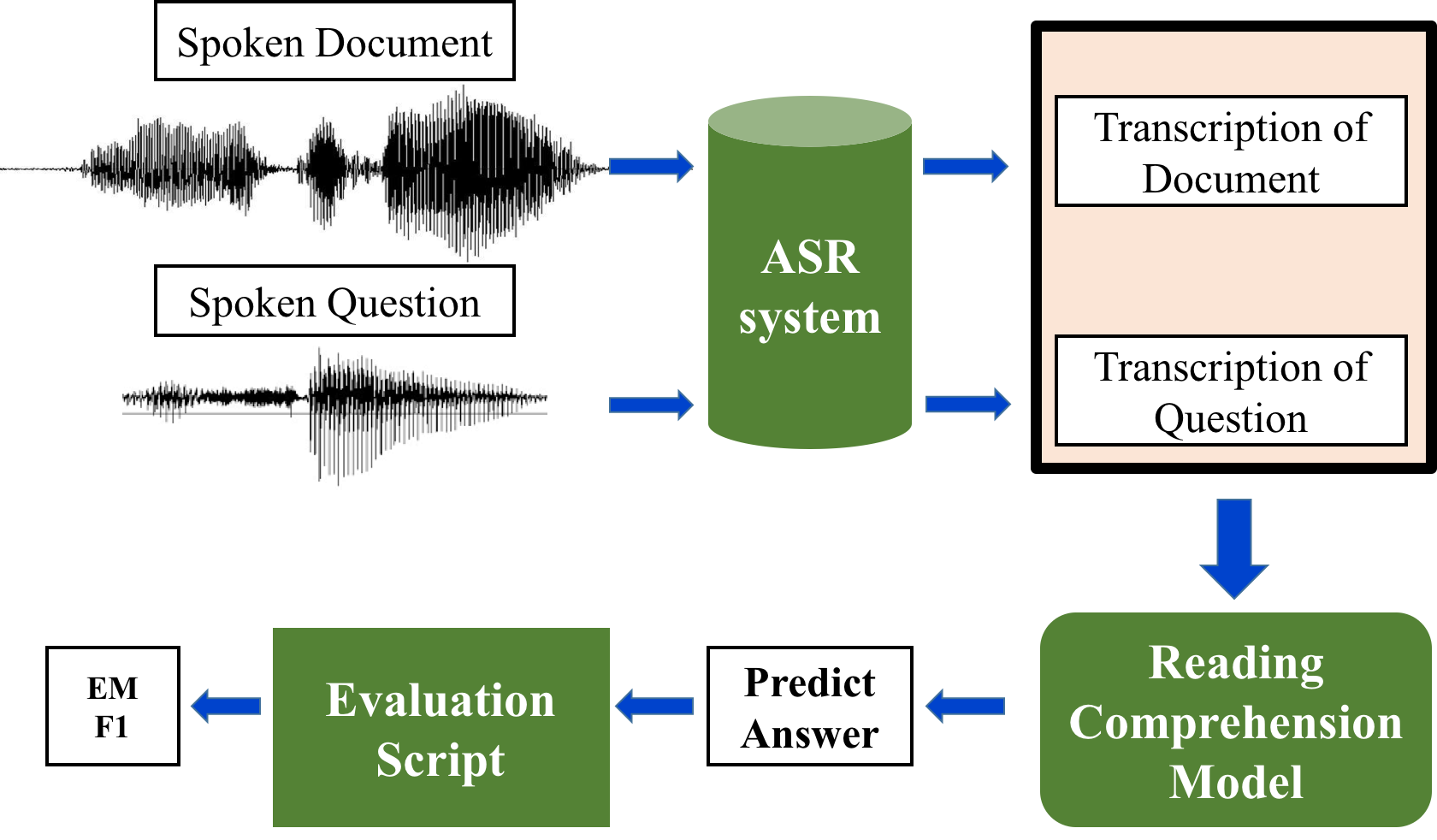}
  \caption{Flow diagram of the SQA system and the standard evaluation method. 
  Given a spoken document and a spoken or text question, an SQA system, which is a  concatenation of ASR module and reading comprehension module, can return a predicted text answer. This  predicted answer is a span in the ASR transcription of the spoken document and will be evaluated by EM/F1 scores.}
  \label{fig:overview}
\end{figure}

In this paper, we introduce a new listening comprehension corpus, Open-Domain Spoken Questions Answering Dataset (ODSQA).
Each example in this dataset is a triple, $(q, d, a)$.
$q$ is a question, which has both text and spoken forms.
$d$ is a multi-sentence spoken-form document.
The  answer $a$ is in text from, which is a word span from the reference transcription of $d$.
%of read Chinese speech named Spoken-Chinese-SQuAD. 
%###########################
An overview architecture of this task is shown in figure~\ref{fig:overview}.
%###########################

%The target language is Chinese, while the source language is English.
%During the testing stage, the documents and questions are all spoken in Chinese.
%For training stage, we have a large amount of English examples, and relatively less Chinese examples.
\subsection{Data Collection}
To build a spoken version QA dataset, we conducted the following procedures to generate spoken documents and questions. 
Our reference texts are from Delta Reading Comprehension Dataset (DRCD), which is an open domain traditional Chinese machine reading comprehension (MRC) dataset~\cite{shao2018drcd}. %Lee: 記得引用台達電放在 arxiv 的文章
Each data example in DRCD is a triple of $(q, d, a)$ in which $q$ is a text-form question, $d$ is a multi-sentence text-form document that contains the answer $a$ as an extraction segment. 
In DRCD, training set contains 26,936 questions with 8,014 paragraphs, the development set contains 3,524 questions with 1,000 paragraphs and the testing set contains 3,485 questions with 1,000 paragraphs. The training set and development set are publicly available, while the testing set is not. 
Therefore, the DRCD training set was used as the reference text of the training set of ODSQA, and the DRCD development set was used as the reference text of the testing set of ODSQA.

20 speakers were recruited to read the questions and paragraphs in the development set of DRCD. %Lee: 請註明與者數量
All the recruited speakers were native Chinese speakers and used Chinese as their primary language. 
For document, each sentence was shown to speaker respectively. 
The speaker was required to speak one sentence at a time. 
All the sentences of the same document were guaranteed to be spoken by the same speaker. 
%To simulate the real life user scenario, for example, an user speaks a question about an already recorded document. 
%We didn't restrict the document and the question from the same data example to be spoken by the same speaker.
Because in the  real-life user scenario, it is more possible that an user enters a spoken question, and machine answers the question based on an already recorded spoken document collection. The document and the question from the same data example do not have to be recorded by the same speakers.
%################################# % 這句應該放到 experimental setup
%If the answer of a question did not exist in the ASR transcriptions of the associated spoken article, we removed  the question-answer pair from the ODSQA dataset because these examples are too difficult for listening comprehension machine at this stage. 
%#################################
%講解蒐集流程

%\subsection{Data Statistics}
We collected 3,654 question answer pairs as the testing set.
The corpus is released\footnote{ODSQA: OPEN-DOMAIN SPOKEN QUESTION ANSWERING DATASET \\
\url{https://github.com/chiahsuan156/ODSQA}}. 
The speech was all sampled at 16 kHz due to its common usage among the speech community, but also because the ASR model we adopted was trained on 16 kHz audio waveforms. 
An example of a corresponding pair between DRCD and ODSQA is shown in column(1) and (2) of Table~\ref{tab:EXAMPLE}. 
The detailed information of ODSQA about the speakers, audio total length and Word Error Rate are listed in row(1) of Table~\ref{tab:statistics}. 

\subsection{Evaluation  Metrics} 
In this task, when the model is given a spoken document, it needs to find the answer of a question from the transcriptions of the spoken document. 
SQA can be solved by the concatenation of ASR module and reading comprehension module.
%The flow diagram of the considered extraction-based spoken question answering task is illustrated in Figure~\ref{fig:overview}. -> 這句話前面有了？
Given a query and  the ASR transcriptions of a spoken document, the reading comprehension module can output a text answer. 
The most intuitive way to evaluate the text answer is to directly compute the \textbf{Exact Match (EM)} and \textbf{Macro-averaged F1 scores(F1)} between the predicted text answer and the ground-truth text answer.
If the predicted text answer and the ground-truth text answer are exactly the same, then the EM score is 1, otherwise 0.
The F1 score is based on the precision and recall. 
Precision is the percentage of Chinese characters in the predicted answer existing in the ground-truth answer, while recall is the percentage of Chinese characters in the ground-truth answer also appearing in the predicted answer. 
The EM and F1 scores of each testing example are averaged to be the final EM and F1 score. 
We used the standard evaluation script from SQuAD\cite{rajpurkar2016squad} to evaluate the performance. 

\begin{comment}
\begin{table*}[]
\caption{An example in ODSQA and the corresponding reference texts in DRCD.The English 
translations were added here for easy reading. Document is denoted as D here.} 
\begin{CJK}{UTF8}{bkai}{}
%\fontsize{8}{8}\selectfont
\label{tab:EXAMPLE}
\begin{tabular}{|c p{4cm}|}
\hline
\multirow{7}{*}{\textbf{Text Document}}  & 
黃河，在中國古代稱作河水，簡稱河，是中國的第二長河，僅次於長江...
“Yellow River, which was called as river water in ancient China, briefly called as river, is the second longest river in China, second only to Yangtze...”
\\
\hline
\multirow{1}{*}{\textbf{Spoken Document}} & xxxxxxx \\
\hline
\multirow{3}{*}{\textbf{Question}} & 中國的第一長河為? “What is the longest river in China?” \\
\hline
\multirow{1}{*}{\textbf{Spoken Question}} & xxxxxxx \\
\hline
\textbf{Ground truth} & 長江 “Yangtze” \\ 
\hline
\end{tabular}
\end{CJK}
\end{table*}
\end{comment}

\begin{center}
\begin{table*}[]
\centering
\caption{An example in ODSQA and the corresponding reference texts in DRCD.The English 
translations were added here for easy reading. } 
\begin{CJK}{UTF8}{bkai}{}
%\fontsize{8}{8}\selectfont
\label{tab:EXAMPLE}
\begin{tabular}{|c || p{3.5cm} |p{3.5cm}| p{3.5cm}|p{3.5cm}|}
\hline
\multirow{1}{*}{\textbf{Data}} & \textbf{(1) DRCD} & \textbf{(2) ODSQA} & \textbf{(3) DRCD-TTS } & \textbf{(4) DRCD-backtrans } \\
\hline
\multirow{7}{*}{\textbf{D}} & 
...廣州屬亞熱帶季風海洋性氣候，氣候濕熱易上火的環境使飲涼茶成為廣州人常年的一個生活習慣。
“Guangzhou has a subtropical monsoon maritime climate. Drinking cool tea has become a daily habit of Guangzhou people for a long time due to the humid and hot environment.” &
...廣州屬亞熱帶季風海洋性氣候，氣候濕熱，易上火的環境\textbf{時，應嚴查}成為廣州人常年的一個生活習慣。 

“Guangzhou has a subtropical monsoon maritime climate. \textbf{When} the environment is hot and humid, \textbf{necessarily strictly examination} has become a daily habit for Guangzhou people for a lone time.”
 & ...廣州屬亞熱帶季風海洋性氣候，氣候，\textbf{是誠意上火的環境適應量}，茶成爲廣州人常年的一個生活習慣。
 
 “Guangzhou has a subtropical monsoon maritime climate. Climate, \textbf{being a sincere and hot humid adaptation to the environment}, tea has become a daily habit for Guangzhou people for a lone time.”
 & ...廣州屬亞熱帶季風海洋性氣候，氣候炎熱潮濕，是廣州人喝茶的共同習慣。
 
 “Guangzhou has a subtropical monsoon maritime climate. The climate is hot and humid, which is a common habit of Guangzhou people when drinking tea.”
\\
\hline
%\multirow{3}{*}{\textbf{Q}} & 喝涼茶成為廣州人的習慣是因為廣州屬於何種氣候？ “Drinking cool tea becomes a habit of people in Canton. It is because of what kind of climate?”  & 喝涼茶成為廣州人的習慣是因為廣州屬於何種氣候？ “Drinking cool tea becomes a habit of people in Canton. It is because of what kind of climate?”  & 喝涼茶成為廣州人的習慣，是因為廣州屬於何種氣候？
%“Drinking cool tea becomes a habit of people in Canton. It is because of what kind of climate?”  & 喝涼茶成為廣州人的習慣是因為廣州屬於何種氣候？“Drinking cool tea becomes a habit of people in Canton. It is because of what kind of climate?”\\
%\hline
%\textbf{A} & 亞熱帶季風海洋性氣候 “subtropical monsoon maritime climate” & 亞熱帶季風海洋性氣候 “subtropical monsoon maritime climate” & 亞熱帶季風海洋性氣候 “subtropical monsoon maritime climate” & 亞熱帶季風海洋性氣候 “subtropical monsoon maritime climate” \\ 
\end{tabular}
\end{CJK}
\vspace{-3mm}
\end{table*}
\end{center}

\section{Proposed Approach}
\begin{figure}[t]
  \centering
  \includegraphics[width=\linewidth]{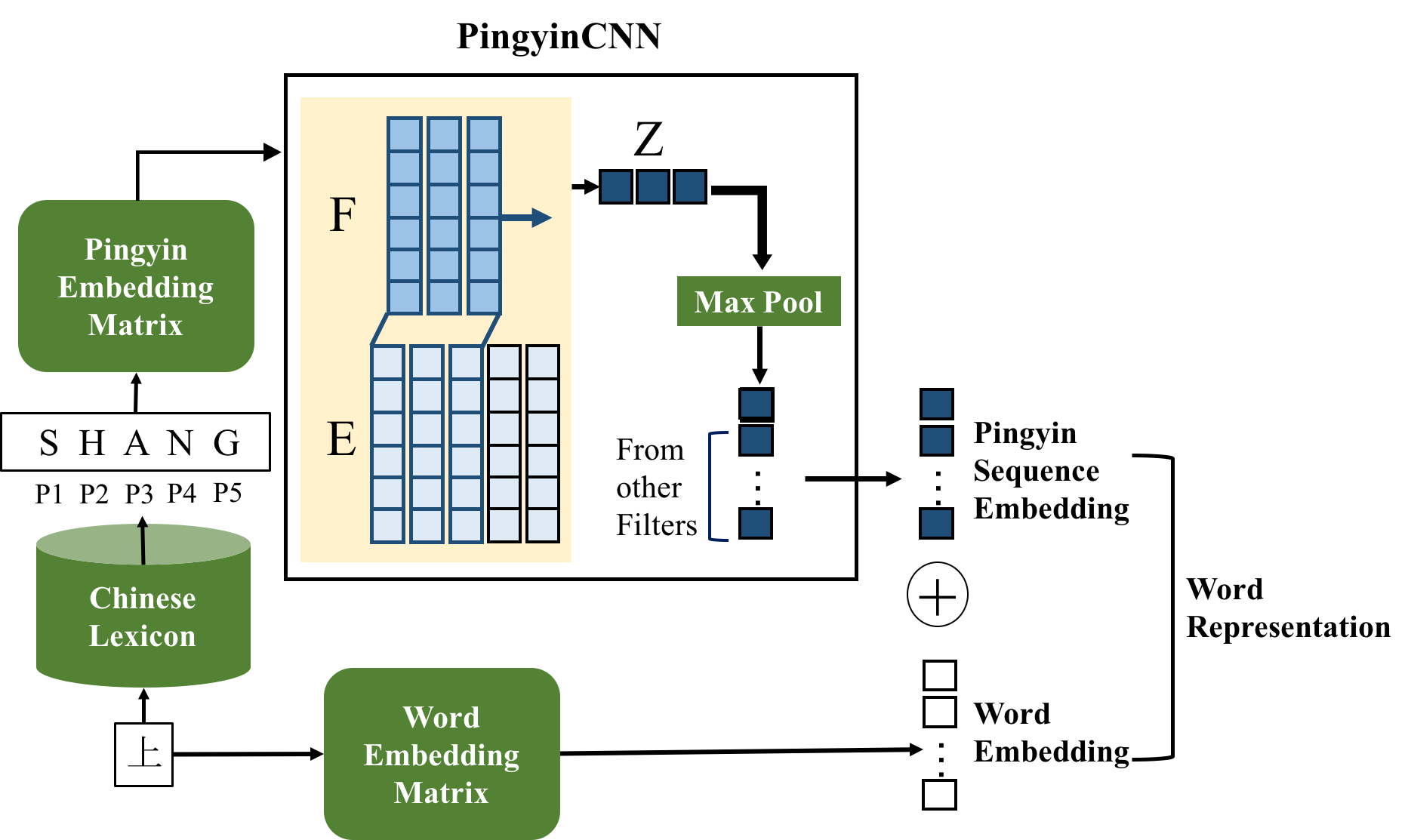}
  \begin{CJK}{UTF8}{bkai}{}
  \caption{Illustration of enhanced word embedding. 
  For a given input word $W$ at the bottom, a sequence of pingyin-tokens $P = [p_1,...,p_l]$ are obtained by looking up in the Chinese lexicon. 
  Each pingyin-token is mapped to a vector $ \mathbb{R}^{d}$ and concatenated to form intermediate matrix $E$.
  $E$ is fed into the 1-D convolutional module. 
  The output $Z$ is further fed into max-pooling layer, and a scalar value is generated. All the scalars from various filters will form the phoneme sequence embedding. 
  Then the pingyin sequence embedding is further concatenated with word embedding as the input  of reading comprehension model.
  In this illustration, the Chinese word 上 (means "up" in English) consists of five pingyin-tokens.}
  %so $E \in \mathbb{R}^{5 \times 6}$, $F \in \mathbb{R}^{3 \times 6}$ and stride is 1.
  \label{fig:phonemeCNN}
  \end{CJK}
\end{figure}

%\subsection{Subword Units}
ASR errors are inevitable, and they can hinder the reasoning of QA models. However, when a transcribed word is wrong, some phonetic sub-word units in the word may still be correctly transcribed. Therefore, building word representation from sub-word units may mitigate the impact of ASR errors.
%\subsubsection{Phoneme as basic component of Word Embedding}
%Previous work \cite{li2016phoneme} applied bi-directional Long Short Term Memory (LSTM) on phoneme sequences to predict acoustic features. \cite{silfverberg2018sound} extracted phoneme embeddings from the embedding layer of an recurrent neural network that is trained to solve a word inflection task. % 這兩篇 paper 的 phoneme embedding 跟本文好像沒甚麼關係？ -- Lee
Pingyin-token sequence of words are used in our experiments. Pinyin, literally meaning “spell out the sound”, is the Romanized phonetic transcription of the Chinese language. 
Each Chinese character consists of one pingyin syllable, and one syllable is composed of a number of pingyin-tokens. 
We adopt one-dimensional Convolution Neural Network (1-D CNN) to generate the word representation from the pingyin-token sequence of a word, and this network is called Pingyin-CNN. 
Our proposed approach is the reminiscent of Char-CNN~\cite{zhang2015character,kim2016character}, which apply 1-D CNN on character sequence to generate distributed representation of word for text classification task. 
Pingyin-CNN is illustrated in Figure~\ref{fig:phonemeCNN}.
We explain how we obtain feature for one word with one filter. %? 
Suppose that a word $W$ consists of a sequence of pingyin-tokens $P = [p_1,...,p_l]$, where $l$ is the number of pingyin-tokens of this word. 
Let $H \in \mathbb{R}^{C \times d}$ be the lookup table pingyin-token embedding matrix, where $C$ is the number of pingyin-tokens, and $d$ is the dimension  of the token embedding. 
In other words, each token corresponds to a $d$-dimensional vector.
Given $P$, after looking up table,  the intermediate token embedding $E \in\mathbb{R}^{l \times d}$ is obtained. 
The convolution between $E$ and a filter $F \in\mathbb{R}^{k \times d}$ is performed with stride 1 to obtain one-dimension vector $Z \in \mathbb{R}^{l-k+1}$. 
After max pooling over $Z$, we obtain a scalar value. 
With a set of  filters, with the above process, we obtain a pingyin-token sequence embedding.
The size of the  pingyin-token sequence embedding is  the number of filters.
%Since we concatenate all the output scalars from different filters, the number of filters determine the size of pingyin-token sequence embedding. 
The filter is essentially scanning pingyin-token n-gram, where the size of n-gram is the height of the filter (the number of $k$ above).
The pingyin-token sequence embedding is concatenated with typical word embedding to obtain new word representation as the input of  reading comprehension model.
All the parameters of filters  and pingyin-token embedding matrix $H$ are end-to-end learned with reading comprehension model. 
%Notice that the word representation obtained from the above architecture can be easily applied to any QA models. In this work, we do not pre-train the phoneme embedding matrix. 
%It is also possible to incorporate other sub-word units like syllable~\cite{luong2013better,botha2014compositional,bian2014knowledge} by the same CNN architecture described above.

%\subsubsection{Syllable as basic component of Word Embedding}
%It  is also possible to incorporate other sub-word units to leverage character level or morphology level information\cite{luong2013better}\cite{botha2014compositional}. 
%Since syllable is a string of concatenated phonemes, we experiment with syllables as basic components for word representation to encode pronunciation signals more effectively. Furthermore, previous works consider syllables can bring morphological knowledge\cite{bian2014knowledge}\cite{choi2017syllable}. We choose the same architecture as Phoneme-CNN to obtain the distributed word representation based on syllables.

%\subsubsection{Word Confidence}
%In Fig.~\ref{fig:phonemeCNN}, we further add the word level confidence score returned by the ASR system as an additional feature. 
%We hope the model learns to put less attention to the words that possess low confidence scores, and improves the performance.
%Word confidence potentially tell the model to put less attention to the words that possess low confidence scores. 

\section{Experiments}
\label{sec:typestyle}
\subsection{Experimental Setup}
\begin{itemize}
  \item \textbf{Speech Recognition}:   We used the  iFLYTEK ASR system \footnote{iFLYTEK ASR system\\\url{ https://www.xfyun.cn/doccenter/asr}} to transcribe both the spoken document and spoken question.
  \item \textbf{Pre-processing}:   We used jieba-TW\footnote{jieba-zh-TW:\\ \url{https://github.com/ldkrsi/jieba-zh-TW}}, a python library specialized for traditional Chinese, to segment sentence into words. The resulting word vocabulary size for DRCD is around 160,000 and the character vocabulary size is around 6,200. We experimented with both words and characters. 
  \item \textbf{Implementation Details}
  
  \textbf{Chinese word embeddings}: We pre-train a Fasttext\cite{bojanowski2016enriching} model on words of traditional Chinese Wikipedia articles\footnote{Wikipedia articles:\\\url{https://dumps.wikimedia.org/zhwiki/}} segmented by jieba-zh-TW. This model can handle Out-of-Vocabulary words with character n-grams. The word embeddings in all our experiments were initialized from this 300 dimensional pre-trained fasttext model and fixed during training for both translated English word and Chinese word. This model is crucial to the performance of the qeustion answering models according to our experimental results.
  
  \textbf{Chinese character embeddings}: We pre-train a skip-gram model on characters of traditional Chinese Wikipedia articles using Gensim\footnote{Genism:\\ \url{ https://radimrehurek.com/gensim/models/word2vec.html} }.
  
  %\textbf{Hyperparameter}:if we add the classification loss in (???) at the beginning of the training process, the shared encoder will have problems to learn useful features for QA. So we linearly increase the hyper-parameter $\lambda$ from 0 to 0.01 in the first 4000 mini-batches to make sure the latent representation of shared encoder became language-independent gradually.
  \end{itemize}
 
  \vspace{-5mm} 

\begin{center}
\begin{table*}[]
\centering
\caption{Data statistics of ODSQA, DRCD-TTS and DRCD-backtrans. The average document length and the average question length are denoted as Avg D Len and Avg Q Len respectively and they stand for the total numbers of Chinese characters. Since training with noisy data and different speakers will make QA model more robust during testing, the number of speakers is large. ODSQA testing set is denoted as ODSQA-test. }
\label{tab:statistics}
\begin{tabular}{|c|c|c|c|c|c|c|c|c|}
 \hline
\textbf{Subsets} &  \textbf{QApair}& \textbf{Hours} &\textbf{M-spkrs}&\textbf{F-Spkrs}&\textbf{WER(\%)}&\textbf{WER-Q(\%)}&\textbf{Avg D Len}&\textbf{Avg Q Len}\\
 \hline
 \hline
 (1) ODSQA-test & 1,465 & 25.28  & 7 & 13 & 19.11 & 18.57 & 428.32 & 22.08\\
\hline
 (2) DRCD-TTS  & 16,746 & -- & -- & -- & 33.63 & -- & 332.80 & 20.53\\
\hline
 (3) DRCD-backtrans & 15,238 & -- & -- & -- & 45.64 & -- & 439.55 & 20.75\\
\hline
\end{tabular}
\vspace{-3mm}
\end{table*}
\end{center}

  \vspace{-8mm}
  
\subsection{Baselines}
\label{subsec:baseline}
We chose several competitive reading comprehension models here. %that have acquired state-of-the-art results on SQuAD. 
The models  are listed as follow:
\begin{itemize}
\item \textbf{BiDirectional Attention Flow (BiDAF)}~\cite{seo2016bidirectional}:   In BIDAF, both character-level and word-level embeddings are incorporated. A Bi-directional attention flow mechanism, which computes attentions in two directions: from context to query as well as from query to
context is introduced to obtain a query-aware context representation.
  \item \textbf{R-NET}~\cite{wang2017gated}:   In R-NET, the dependency in long context is captured more than plain recurrent neural network. A self-matching mechanism is introduced to dynamically refine context representation with information from the whole context.
  \item \textbf{QANet}~\cite{yu2018qanet}:   There are completely no recurrent networks in QANet. Its encoder is composed of exlusively of convolution and self-attention. The intuition is that convolution components model local interactions and self-attention components model global interactions. Due to the removal of recurrent networks, it's training speed is 5x faster than BiDAF when reaching the same performance on SQuAD dataset.
  \item \textbf{FusionNet}~\cite{huang2017fusionnet}:   There are mainly two contributions in FusionNet.  First is the \textbf{History of Word}, in which all representations of a word from lowest level word embedding to the highest semantic level are concatenated to be the final representation of this word. Second is the \textbf{Fully-aware Multi-level Attention Mechanism}, which captures the complete information in one text (such as a question) and exploits it in its counterpart (such as a context or passage) layer by layer.
  \item \textbf{Dr.QA}~\cite{chen2017reading}:   Dr.QA is a rather simple neural network architecture compared to the previous introduced models. It basically is composed of multi-layer bidirectional long short-term memory networks. It utilizes some linguistic features such as part-of-speech tagging and name entity recognition. % Because of our setting, we didn't utilize these two features.
  \end{itemize}
In our task, during testing stage, all the baseline QA models take into a machine-transcribed spoken document and a machine-transcribed spoken question as input, and the output is an extracted span from the ASR transcription of document. 
We train these baseline QA models on DRCD training set and compare the performance between DRCD dev set and ODSQA testing set.

\subsection{Artificially Generated Corpus}
It is reported that training on transcriptions with ASR errors are better than training on text.\cite{li2018spoken}, so we conduct the following procedures to generate transcriptions of spoken version DRCD. First, we used iFLYTEK Text-to-Speech system \footnote{iFLYTEK Text-to-Speech system\\
\url{ https://www.xfyun.cn/doccenter/tts}}
to generate the spoken version of the articles in DRCD. Then we utilized iFLYTEK ASR system to obtain the corresponding ASR transcriptions. In this corpus, we left the questions in the text form. If the answer to a question did not exist in the ASR transcriptions of the associated article, we removed the question-answer pair from the corpus. This artificially generated corpus is called \textbf{DRCD-TTS} and its data statistics is shown in row(2) of Table~\ref{tab:statistics}.

\subsection{Back-translation Corpus}
To improve the robustness to speech recognition errors of QA model, we augmented DRCD training dataset with back-translation. We conduct the following procedures to generate an augmented training set. First, the DRCD training set is translated using Google Translation system into English. Then this set is translated back into Chinese through Google Translation system. We chose English as the pivot language here because English is the most common language and the translation quality is probably the best. Because the task is extraction-based QA, the ground-truth answer must exist in the document.
Therefore, we removed the examples which cannot fulfill the requirement of extraction-based QA after translation. This resulting dataset is called \textbf{DRCD-backtrans} and its statistics is shown in row(3) of Table~\ref{tab:statistics}.

\begin{comment}
\begin{figure}[t]
  \centering
  \includegraphics[width=\linewidth]{augmentation.png}
  \caption{Flow diagram of the generation procedure of DRCD-backtrans and DRCD-TTS dataset.}
  \label{fig:augmentation}
\end{figure}
\end{comment}

\subsection{Result}
First of all, we show the performance of the baseline models that are briefly introduced in subsection 5.2. All the QA models were trained on DRCD then test on DRCD dev set and ODSQA testing set respectively to compare the performance between text documents and spoken documents. 

%\footnote{All of the QA models take in both character-level and word-level embeddings, but we only used word-level embeddings in the word-base settings and character-level embeddings in the character-base settings. Furthermore, DrQA originally incorporated POS and NER tagging, we chose not to use them to make the comparison fair.}

Secondly, we compare the performance of QA models with and without the proposed pingyin sequence embedding. 
Thirdly, we show how co-training with \textbf{DRCD-TTS} and \textbf{DRCD-backtrans} benefit. 
Last but not the least, we compare the performance between spoken question and text question.
%Finally, we experimented with our proposed approach including the ablation studies of shared encoder, adversarial learning and Chinese Syllables.  

\textbf{Investigating the Impact of ASR Errors}.
We trained five reading comprehension models mentioned in Section 5.2 on the DRCD training set and these five models are tested on DRCD dev set and ODSQA testing set. 
%The number of question answer pairs in the ODSQA testing set is less than the original DRCD dev set because some of the examples are removed as mentioned in Section 3.2. 
%If the answer of a question did not exist in the ASR transcriptions of the associated spoken article, we removed  the question-answer pair from the ODSQA dataset because these examples are too difficult for listening comprehension machine at this stage. 
In the following experiments, we do not consider the spoken documents whose answers do not exist in the ASR transcriptions because the model can never obtain the correct answers in these cases. %
To make the comparison fair, the DRCD dev set are filtered to contain only the same set of examples. %that are also in ODSQA testing set. 
As shown in Table~\ref{tab:stateoftheart}, across the five models with char-based input, the average F1 score on the text document is 81.05\%.
The average F1 score fell to 63.67\%  when there are ASR errors. Similar phenomenon is observed on EM.
The impact of ASR errors is significant for machine comprehension models. 
Since the author of BiDAF released its source code\footnote{BiDAF:Bi-directional Attention Flow for Machine Comprehension
\\\url{https://github.com/allenai/bi-att-flow}} and its decent performance over ODSQA testing set, we use it as the base model with char-based input for further experiments. 

\begin{table}[]
\centering
\vspace{-3mm}
\caption{Experiment results for state-of-the-art QA models demonstrating degrading performance under spoken data. All models were trained on the full DRCD training set. FusionNet is denoted by F-NET. DRCD dev set and ODSQA testing set are denoted by DRCD-dev and ODSQA-test, respectively. }
\label{tab:stateoftheart}
\begin{tabular}{|c|c|c|c|c|}
\hline
\multicolumn{1}{|c|}{\multirow{2}{*}{\textbf{MODEL}}} &
\multicolumn{2}{|c|}{\textbf{DRCD-dev}} & \multicolumn{2}{|c|}{\textbf{ODSQA-test}}  \\
 \cline{2-3}\cline{4-5}
\multicolumn{1}{|c|}{} & EM & F1 & EM & F1 \\
\hline
\hline
BiDAF-word(a) & 56.45 & 70.57 & 39.38 & 55.1   \\
BiDAF-char(b) & 70.23 & 81.65 & 55.29 & 67.16     \\
\hline
R-NET-word  & 70.38 & 79.25 & 36.68 & 46.55 \\
R-NET-char & 69.90 & 79.49 & 43.44 & 55.83   \\
\hline
QAnet-word & 69.83 & 78.33 &  49.80 & 59.35   \\
QAnet-char & 70.78 & 80.83 & 46.52 & 59.11   \\
\hline
Dr.QA-word & 63.21 & 74.11 & 41.39 & 54.28 \\
Dr.QA-char & 70.24 & 81.19 & 56.22 & 68.99  \\
\hline
F-Net-word & 57.54 & 70.86 & 45.39 & 57.40 \\
F-Net-char & 71.33 & 82.12 & 47.98 & 67.26   \\
\hline
\hline
\textbf{Average-word} & 63.48 & 74.62 & 42.52 & 54.53 \\
\textbf{Average-char} & 70.49 & 81.05 & 49.89 & 63.67   \\
\hline
\end{tabular}
\vspace{-3mm}
\end{table}

\textbf{Mitigating ASR errors by Subword Units}.
We utilized an open-sourced chinese mandarin lexicon tool \footnote{DaCiDian : an open-sourced chinese mandarin lexicon for automatic speech recognition(ASR) \\ \url{https://github.com/aishell-foundation/DaCiDian}} to convert each word into sequence of pingyin-tokens and then fed the ping-yin tokens into Pingyin-CNN network to obtain pingyin-token sequence embedding. 
In this work, we didn't utilize tone information in pingyin-tokens. We leave it as a future work.
The network details are listed as follow: pingyin-token embedding size 6, filter size 3x6 and numbers of filters 100.  Different from~\cite{li2016phoneme} using one-hot vector, we choose distributed representation vectors to represent   sub-word units.
The experimental results with and without the proposed sub-word unit approach are in Table \ref{tab:approaches}. We see from Table \ref{tab:approaches}, using the combination of word embedding and the proposed pingyin sequence embedding is better than just word embedding (row (b)(d)(f)(h)(j)(l) vs. (a)(c)(e)(g)(i)(k)). The average EM score is improved by 1.3 by using pingyin sequence embedding over ODSQA testing set. 

\textbf{Data augmentation}.
To improve the robustness to speech recognition errors of QA models, we augmented training data DRCD with DRCD-TTS and DRCD-backtrans. The experiment results are shown in Table \ref{tab:approaches}. We can see from Table \ref{tab:approaches}, training with the combination of DRCD and DRCD-backtrans or training with the combination of DRCD and DRCD-TTS are all better than training with only DRCD (row (g)(i) vs. (a) and row(h)(j) vs. row(b)). And finally training with the combination of DRCD, DRCD-TTS and DRCD-backtrans with pingyin sequence embedding obtains the best results (row(l)) which is better than baseline (row (a)) by almost 4 F1 score. Therefore, data augmentation proves to be helpful in boosting performance.

\begin{table}[]
\centering
\vspace{-5mm}
\caption{Comparison experiments demonstrating that the proposed sub-word units improved EM/F1 scores over both DRCD-dev and ODSQA-test. We use BiDAF as our base model  in all experiments. Furthermore, augmenting DRCD with DRCD-TTS and DRCD-backtrans also gain improvements. Training with the combination of DRCD and DRCD-backtrans, the combination of DRCD and DRCD-TTS and the combination of DRCD, DRCD-TTS and DRCD-bakctrans are denoted as DRCD+back, DRCD+TTS and DRCD+TTS+back respectively.}
\label{tab:approaches}
\begin{tabular}{|c|c|c|c|c|}
\hline
\multicolumn{1}{|c|}{\multirow{2}{*}{\textbf{MODEL}}} &
\multicolumn{2}{|c|}{\textbf{DRCD-dev}} & \multicolumn{2}{|c|}{\textbf{ODSQA-test}}  \\
 \cline{2-3}\cline{4-5}
\multicolumn{1}{|c|}{} & EM & F1 & EM & F1 \\
\hline
\hline
DRCD (a) & 70.23 & 81.65 & 55.29 & 67.16   \\
+pingyin (b) & 71.05 & 81.82 & 55.49 & 68.79   \\
\hline
DRCD-TTS (c) & 59.24 & 72.64 & 50.64 & 63.65   \\
+pingyin (d) & 61.36 & 74.22 & 51.74 & 64.59   \\
\hline
DRCD-back (e)  & 58.56 & 72.31 &  46.55 & 61.52   \\
+pingyin (f) & 58.63 & 72.97 & 48.2 & 62.82    \\
\hline
\hline
DRCD+TTS (i) & 70.51 & 81.85 & 55.97 & 69.31   \\
+pingyin (j) & 71.53 & 82.42 & 56.65 & 69.45   \\
\hline
DRCD+back (g) & 71.39 & 82.28 & 55.29 & 68.49   \\
+pingyin (h) & 71.8 & 82.4 & 57.6 & 69.26   \\
\hline
\hline
DRCD+TTS+back (k) & 72.21 & 82.8 & 57.61 & 70.29   \\
+pingyin (l) & \textbf{72.76} & \textbf{83.15} & \textbf{59.52} & \textbf{71.01} \\
\hline
\hline
\textbf{Average (m)} & 67.02 & 78.92 & 53.55 & 66.73 \\
\textbf{Average-pingyin (n)} & 67.85 & 79.49 & 54.86 & 67.65 \\
\hline
\end{tabular}
\vspace{-5mm}
\end{table}

\textbf{Comparison Between Text Question and ASR Transcribed Question}.
ASR errors on question will affect the reasoning of a QA model. In this part, we compare the performance between input with text questions and input with ASR-transcribed questions. We can see from Table \ref{tab:textq}, the average F1 score fell from 71.61\%  to 66.73\% when there are ASR errors in question. Similar phenomenon is observed on EM. Once again, we can see that using pingyin sequence embedding brings improvement (row(h) vs (g)) even with text question as input.

\begin{table}[]
\centering
\vspace{-8mm}
\caption{Comparison experiments between input with text question and input with transcribed question. We use BiDAF as our base model in all experiments.}
\label{tab:textq}
\begin{tabular}{|c|c|c|c|c|}
\hline
\multicolumn{1}{|c|}{\multirow{2}{*}{\textbf{MODEL}}} &
\multicolumn{2}{|c|}{\textbf{Text-Q}} & \multicolumn{2}{|c|}{\textbf{Spoken-Q}}  \\
 \cline{2-3}\cline{4-5}
\multicolumn{1}{|c|}{} & EM & F1 & EM & F1 \\
\hline
\hline
DRCD (a) & 59.63 & 72.02 & 55.29 & 67.16   \\
+pingyin (b) & 61.47 & 72.93 & 55.49 & 68.79   \\
\hline
DRCD-TTS (c) & 54.43 & 67.18 & 50.64 & 63.65   \\
+pingyin (d) & 55.39 & 68.12 & 51.74 & 64.59   \\
\hline
DRCD-back (a)  & 52.45 & 67.13 &  46.55 & 61.52   \\
+pingyin (b) & 53.41 & 68.57 & 48.2 & 62.82    \\
\hline
\hline
DRCD+TTS (i) & 61.95 & 73.78 & 55.97 & 69.31   \\
+pingyin (j) & 62.43 & 74.3 & 56.65 & 69.45   \\
\hline
DRCD+back (c) & 62.22 & 74.33 & 55.29 & 68.49   \\
+pingyin (d) & 62.7 & 74.81 & 57.6 & 69.26   \\
\hline
\hline
DRCD+TTS+back (e) & 63.11 & 75.27 & 58.29 & 69.94   \\
+pingyin (f) & 64.54 & 75.63 & 59.52 & 70.95 \\
\hline
\hline
\textbf{Average (g)} & 58.96 & 71.61 & 53.55 & 66.73\\
\textbf{Average-pingyin (h)} & 59.99 & 72.39 & 54.86 & 67.65 \\
\hline
\end{tabular}
\vspace{-5mm}
\end{table}

\begin{comment}
\begin{table}[]
\centering
\caption{EM/F1 scores over ODSQA testing set. Row (a) is the word-based baseline BIDAF model trained on DRCD training set. All the proposed approaches were trained jointly using the training examples of DRCD and DRCD-TTS. }
\label{tab:approaches}
\begin{tabular}{c|c|c|c|c}
\hline
\multicolumn{2}{|c|}{{\textbf{Approaches}}} & \multicolumn{1}{|c|}{} &\textbf{EM} & \textbf{F1}  \\
\hline
\hline
\multicolumn{2}{|c|}{{Baseline}} & (a)  & 56.27 & 74.49 \\
\hline\hline

CHAR & share & (f) & 00 & 00 \\
CHAR & share +syllable & (g) & 00 & 00 \\
CHAR & share + GAN & (h) & 00 & 00 \\
CHAR & share +syllable+ GAN & (i) & 00 & 00 \\
\hline
%\multicolumn{2}{|c|}{{2 context embed}} & (f) & \textbf{46.96} & \textbf{64.75} \\
%\multicolumn{2}{|c|}{{2 context embed+WGAN}} & (g) & \textbf{50.47} & \textbf{68.15} \\
%SHARE+WGAN & (h) & \textbf{} & \textbf{} \\
\end{tabular}
\end{table}
\end{comment}

\section{Concluding Remarks}
\vspace{-1.5mm}
We release an SQA dataset, ODSQA. 
By testing several  models, we found that ASR errors have catastrophic impact on SQA.
We found that subword units bring consistent improvements over all the models.
We also found that using back-translation and TTS to augment the text-based QA training examples can help SQA.
In the future work, we are collecting more data for SQA.

%Average EM score improves 1.3 by using pingyin sequence embedding.
%By the combination of our proposed data augmentation approach and phonetic sub-word unit approach, 4 F1 score improvement is observed. 

% Below is an example of how to insert images. Delete the ``\vspace'' line,
% uncomment the preceding line ``\centerline...'' and replace ``imageX.ps''
% with a suitable PostScript file name.
% -------------------------------------------------------------------------

%\vfill
%\pagebreak

\bibliographystyle{IEEEbib}
\bibliography{strings,refs}

\end{document}